\documentclass[nohyperref]{article}

\usepackage{microtype}
\usepackage{graphicx}
\usepackage{subfigure}
\usepackage{booktabs} 
\usepackage{array}

\usepackage{siunitx, tabularx, etoolbox, booktabs}
\usepackage{hyperref}

\usepackage[accepted]{icml2022_pods}

\usepackage{amsmath}
\usepackage{amssymb}
\usepackage{mathtools}
\usepackage{amsthm}

\newcommand{\rn}[1]{\mathbb{R}^{#1}}
\newcommand{\rz}[0]{\mathbb{R}}

\usepackage{physics}

\usepackage[capitalize,noabbrev]{cleveref}

\theoremstyle{plain}
\newtheorem{theorem}{Theorem}[section]

\newtheorem{lemma}[theorem]{Lemma}

\theoremstyle{definition}

\theoremstyle{remark}

\icmltitlerunning{Time Series Prediction under Distribution Shift using Differentiable Forgetting}

\begin{document}

\twocolumn[
\icmltitle{Time Series Prediction under Distribution Shift using Differentiable Forgetting}




\begin{icmlauthorlist}
\icmlauthor{Stefanos Bennett}{dpt,comp}
\icmlauthor{Jase Clarkson}{dpt,comp}
\end{icmlauthorlist}

\icmlaffiliation{dpt}{Department of Statistics, University of Oxford, UK}
\icmlaffiliation{comp}{Alan Turing Institute, London, UK}

\icmlcorrespondingauthor{Stefanos Bennett}{stefanos.bennett@stats.ox.ac.uk}
\icmlcorrespondingauthor{Jase Clarkson}{jason.clarkson@stats.ox.ac.uk}

\icmlkeywords{Time Series, Forecasting, Distribution Shift, Hyperparameter Optimisation}

\vskip 0.3in
]



\printAffiliationsAndNotice{}  

\begin{abstract}
    Time series prediction is often complicated by
    distribution shift which demands adaptive models to accommodate time-varying distributions. We frame time series prediction under distribution shift as a weighted empirical risk minimisation problem. The weighting of previous observations in the empirical risk is determined by a forgetting mechanism which controls the trade-off between the relevancy and effective sample size that is used for the estimation of the predictive model. In contrast to previous work, we propose a gradient-based learning method for the parameters of the forgetting mechanism. This speeds up optimisation and therefore allows more expressive forgetting mechanisms.
\end{abstract}

\section{Introduction}
We introduce a new framework for the problem of predicting time series that exhibit distribution shift, also known as \textit{concept drift} in the domain of engineering \cite{Lu2019} or \textit{forecasting under unstable environments} in economics \cite{Rossi2012}. Time series exhibiting distribution shift appear in a wide range of domains \cite{Zliobaite16} from industrial management to biomedical applications. Financial time series distribution shift has been studied in the concept drift literature \cite{Harries96}, as well as more recently in the statistics \cite{McCarthy2016} and learning theory \cite{Kuznetsov2020} literatures.
We develop the forgetting mechanism approach to predicting time series under distribution shift. The key novel contribution of our work is a gradient-based learning framework for directly optimising the validation performance of differentiable forgetting mechanisms. The main advantages of our framework are that it is easy-to-use with any machine learning method based on empirical risk minimisation, provides quick estimation of forgetting mechanism hyperparameters -- thereby facilitating expressive forgetting mechanisms -- and maintains competitive performance to other common approaches for handling time series distribution shift. While our framework can be applied to any general time series distribution shift problem, we show how it may in particular improve financial risk predictive modelling. Our code is available \href{https://github.com/jase-clarkson/pods_2022_icml_ts}{here}\footnote{\url{https://github.com/jase-clarkson/pods_2022_icml_ts}}.


\section{Problem Definition} \label{sec:Problem Definition}
Consider a sequential supervised learning problem where for each $t = 1, \ldots, T$, we wish to learn a mapping $f(\, \cdot \, ; \theta_t) : X_t \mapsto Y_t$ between features $X_t$ and labels $Y_t$ where $\theta_t \in \Theta$, for some parameter space $\Theta$. The labels, conditional on the features, follow a time-varying distribution $Y_t | X \sim \pi_t( \cdot | X)$. At each time point $t - 1$, we are interested in minimising the one-step-ahead path-dependent risk $R_{t}(\theta) = \mathbb{E}_{Y_t \sim \pi_t( \cdot | X_t)} \left[ L\left( f(X_t; \theta), Y_t \right) | \{(X_{\tau}, Y_\tau)\}_{\tau=1}^{t-1} \right]$, over $\theta \in \Theta$, for a given loss function $L$. We follow an Empirical Risk Minimisation (ERM) \cite{Kuznetsov2020} approach in which we estimate the optimal model parameters $\theta_t$ by minimising the risk estimator $\hat{R}_{t}(\theta)$. 

\section{Model Description} \label{sec:Model Description}

\subsection{Model Adaptation to Distribution Shift via Weighted Empirical Risk Minimisation} \label{sec:Model Adaptation to Distribution Shift via Weighted Empirical Risk Minimisation}
Our approach to accommodate distribution shift in the ERM formulation is through the choice of the one-step-ahead risk estimator 

\vspace{-5mm}
\begin{equation*}
\hat{R}_{t}(\theta) = \sum_{\tau=1}^{t - 1} \alpha_{| t - 1 - \tau |} L\left( f(X_{\tau}; \theta), Y_{\tau} \right)%
\end{equation*}
\vspace{-3mm}  

where the weights $\{ \alpha_i \}_{i=0}^{T - 1}$ are the outputs of a forgetting mechanism $\alpha(i; \eta) = \alpha_i, \, i = 0, \ldots, \, T-1$. 
\begin{figure}
    \centering
    \includegraphics[width=0.48\textwidth,trim={0 10 0 0},clip]{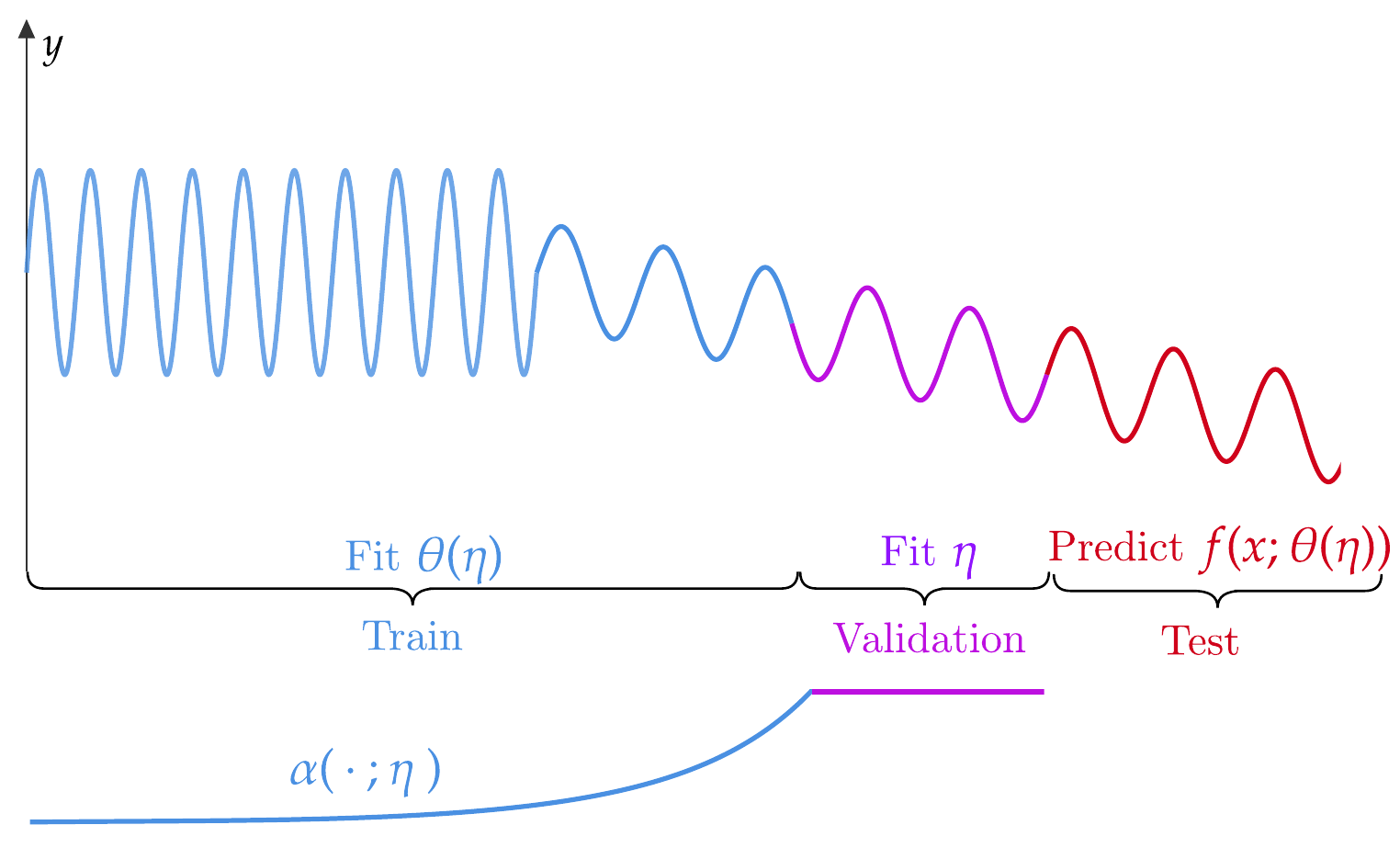}
    \vspace{-0.6cm}
    \caption{Method visualisation: time series training samples (top) are weighted by the forgetting mechanism $\alpha(\cdot; \eta)$ (bottom). The forgetting mechanism, which is parameterised by $\eta$ and optimises predictive performance on the validation set data, assigns weight on the most recent, representative training samples.}
    \label{fig:method}
\end{figure}
The purpose of re-weighting empirical loss samples is to find a linear combination of training set sample losses that yields a model with improved generalisation. We can encode inductive biases by choosing the functional form of the forgetting mechanism.
For instance, a monotonically decreasing forgetting mechanism encodes the hypothesis that more recent samples are more relevant for training. We introduce examples of specific functional forms in Section \ref{sec:Experiments - Methods}. The learning of forgetting mechanism parameters $\eta$ controls the trade-off between the adaptivity of the forecaster and the statistical efficiency of the learning procedure: forgetting mechanisms that assign high weight to a few points are able to capture more relevant data at the cost of reducing the effective sample size that is used to fit the underlying model. 

\vspace{-1mm} 
\subsection{Differentiable Bi-Level Optimisation
} \label{sec:Differentiable Bi-Level Optimisation for Weighted Empirical Risk Minimisation}
 
We jointly learn the predictive model and forgetting mechanism parameters $(\theta, \eta)$ by solving the bi-level optimisation problem
\begin{align*}
    \min_{\eta} \; g^{U}(\eta,\hat{\theta}) \;
    \textrm{\textit{such that}} \; \hat{\theta} \in \underset{\theta}{\mathrm{argmin}} \; g^L (\eta,\theta),  \label{eq:lowerargmin}
\end{align*}
where $g^U$ and $g^L$ represent the upper and lower level objective functions
\begin{align*}
    g^U(\eta, \theta) &:= \sum_{(X_t, Y_t) \in D_{valid}} L(f(X_t; \theta), Y_t) \\
    g^L(\eta, \theta) &:= \sum_{(X_{\tau}, Y_{\tau}) \in D_{train}} \alpha(t^* - \tau; \eta) L(f(X_{\tau}; \theta), Y_{\tau}).
\end{align*}
Here, $D_{train}$ and $D_{valid}$ denote training and validation sets, where the validation set is selected to be the most recent $T - t^*$ data samples, for some choice of $t^*$. The assumption underlying this choice of validation set is that the distribution $\pi_t( \cdot | X)$ does not, on average, substantially change in any given small time span \cite{Kuznetsov2020}. 

Applying the implicit function theorem to bi-level optimisation, we are able to perform gradient-based learning of the forgetting mechanism parameters $\eta$. 
Specifically, \citet{Gould2016} provide the following Lemma for differentiating through the argmin in the lower level problem: 
\begin{lemma} \label{lemma:bi-level_optimisation}
Let $g^L: \rn{m} \times \rn{n} \rightarrow \rz$ be an element of $C^2(\rn{m}, \rn{n})$, the set of twice continuously differentiable functions in each argument, and let $\hat{\theta}(\eta) = \underset{\theta}{\textrm{argmin}} \, g^L(\eta,\theta)$. Then the derivative of $\hat{\theta}$ with respect to $\eta_i, \, i=1, \ldots, m$ is given by 
\begin{equation} \label{eq:theta_alpha_deriv_alpha}
    \pdv{\hat{\theta}(\eta)}{\eta_i} = 
    -\left(\left. \nabla^2_{\theta}{g^L(\eta, \theta)} \right|_{\hat{\theta}} \right)^{-1}
    \pdv{}{\eta_i} \left. \nabla_{\theta} g^L(\eta, \theta) \right |_{\hat{\theta}}.
\end{equation}
\end{lemma}
Therefore, by the chain rule, the total derivative of $g^U(\eta, \hat{\theta}(\eta))$ with respect to $\eta_i$ is given by: 
\begin{equation} \label{eq:total_deriv_alpha}
    \frac{d}{d \eta_i} g^U(\eta, \hat{\theta}(\eta)) = 
    \pdv{g^U}{\eta_i} + 
    \left( \nabla_{\theta} g^U \right)^T \left. \pdv{\hat{\theta}(\eta)}{\eta_i}
    \right|_{\hat{\theta}}
\end{equation}
where we use the gradient expression in equation \eqref{eq:theta_alpha_deriv_alpha}. Equation \eqref{eq:total_deriv_alpha} gives us the required expression 
to apply gradient descent in $\eta$ to the upper-level objective function:
\begin{equation*}
    \eta \leftarrow \eta 
    - \lambda \frac{d}{d \eta} g^U(\eta, \boldsymbol{\hat{\theta}}(\eta)). 
\end{equation*}
where $\lambda \in \rz$ is the step size.

\subsection{Gradient Optimisation Routine} \label{sec:Gradient Optimisation Routine}
Since the upper level problem is generally non-convex with multiple local minima, we propose running gradient descent with multiple restarts to explore different local solutions. 

Once we have estimated the parameters $\eta$ using bi-level optimisation on the train-validation split, we refit our predictive model on the combined train and validation sets. The weighting scheme for the training data points is given by the forgetting mechanism using the estimated $\eta$ while a uniform weighting scheme is applied to the validation data points: each of these is given the same weight as the last training set point, $\alpha(0, \hat{\eta})$. 

There are further considerations when applying the bi-level optimisation procedure using a neural network as the underlying predictive model. Optimisation techniques such as flatness-aware learning \cite{SWA} and Hessian approximation \cite{duv_hyp} are relevant for numerical stability and large Hessian computations; see Appendix \ref{sec: Appendix - Bi-Level Optimisation for Deep Learning} for further discussion.

\vspace{-2mm} 
\section{Related Work} \label{sec:Related Work}
Recursive least squares \cite{Haykin86, Rossi2012, Harvey90} provides an early example of the use of a simple forgetting mechanism for model adaptation. Since then, the concept drift literature has studied the use of forgetting mechanisms for time series distribution shift on an algorithm-by-algorithm basis \cite{Koychev2000, Klinkenberg2004}. 
The field of non-stationary time series forecasting \cite{McCarthy2016,Kuznetsov2020} has also considered the use of forgetting mechanisms for dealing with distribution shift. \citet{McCarthy2016} propose an alternative Bayesian formulation to our ERM-based approach. \citet{Kuznetsov2020} propose an estimator for the one-step-ahead risk that is based on a weighted empirical risk. They derive a learning guarantee on the one-step-ahead generalisation error that holds in the non-stationary time series setting. They minimise this learning guarantee over the weights in their empirical risk and the parameters of their model. 
The first key difference between their work and ours is that we model the weights using a  parametric forgetting mechanism. Secondly, they optimise an upper bound on the generalisation error to derive $\{ \alpha_i \}$ estimates whereas we directly minimise the validation-set prediction error to derive $\{ \alpha_i \}$.

Our work distinguishes itself from the existing concept drift and non-stationary time series forecasting literature as it provides gradient-based learning for a generic forgetting mechanism and ERM-based learning model. A key novelty of our work is gradient-based minimisation of the validation error over the parameters of the forgetting mechanism. Gradient-based approaches are known to be much faster \cite{Hutter18} than grid search, which has typically been used to estimate the forgetting mechanism parameters in the concept drift \cite{Koychev2000, Klinkenberg2004} and non-stationary time series literature \cite{McCarthy2016}. Faster learning allows for more complex, richly-parameterised forgetting mechanisms that would be too computationally burdensome to fit using grid search. Our method connects time series prediction under distribution shift to the growing field of automated machine learning \cite{Hutter18}, allowing us to leverage powerful modern automatic differentiation libraries for efficient implementation.  
 
Beyond time series prediction, the idea of re-weighting empirical loss to address distribution shift has been considered in the context of importance weighting for transfer learning \cite{Fang2020} and robust deep learning \cite{Shu2019}. For instance, in order to address the problem of data noise and class imbalance, \citet{Shu2019} propose to train a neural network that outputs the ERM importance weight of a training sample given that sample's loss. The work of \citet{Shu2019} and \citet{Jenni2018} illustrates the interest in applying differentiable bi-level optimisation methods to the problem of generalisation in non-temporal data settings.

\begin{table*}[htbp] 
\caption{This table reports the Mean Squared Error (MSE) for each model averaged across Monte Carlo repetitions or across time series (for the real data). For each data set, the MSE of the best performing model is shown in bold. A star is placed by a model's MSE whenever its MSE is significantly larger, at the 5\% level, than the best model's MSE using a paired Wilcoxon signed-rank test\footnotemark[3] \cite{Wilcoxon1945}.}
\label{tab:results}
\begin{center}
\robustify\bfseries
\robustify\itshape
\begin{small}
\begin{sc}
\makeatother
\begin{tabular}{|lcccccc|}
\hline
Model & \shortstack{FixedRegime \\($\times 10^{-3}$)} & \shortstack{RandomWalk  \\ ($\times 10^{-3}$)} & \shortstack{RandomRegime \\ ($\times 10^{-3}$)}  & \shortstack{Stat \\ ($\times 10^{-3}$)} & \shortstack{Factor \\ ($\times 10^{-4}$)} & \shortstack{Vol \\ ($\times 10^{-5}$)} \\ 
\hline

Stationary &  {4.00*}   &   {17.2*}\phantom{*}   &   \bfseries 4.20\phantom{*}   & \bfseries 2.54\phantom{*} &    {2.60*}    &      8.77\phantom{*}      \\ 
Window    &   2.62\phantom{*} &  {\phantom{*}3.10*}  &   {4.63*}   & {2.57*}       &  {2.62*}   &    8.96\phantom{*}        \\ 
StateSpace       &  {2.73*}    &   {\phantom{*}3.30*}  & {5.25*}  & {2.60*}      &   {3.81*}     &    {9.75*}     \\ 
Arima            &  {2.65*}    &   {13.4*}\phantom{*}   &   {4.41*}   & {2.57*}       &    {--}    &      {9.24*}      \\ 
Dbf            &    {4.44*}  & {23.3*}\phantom{*} &  {4.55*}   &  {2.64*}      &   {5.11*}    &       {10.7*}     \\ 
GridSearchExp     &   2.63\phantom{*}    &   3.00   &   {4.31*}   & {2.58*}      &   {2.59*}     &      8.94\phantom{*}      \\ 
GradExp     &   {3.96*}    &   {17.2*}\phantom{0}   &   \bfseries 4.20\phantom{*}   & 2.55\phantom{*}     &    {2.59*}  & 8.77\phantom{*}   \\ 
GradMixedDecay   &  \bfseries 2.60\phantom{*}    &   \bfseries 2.80   &   {4.39*}  & {2.57*} & \bfseries 2.49\phantom{*} & \bfseries 8.76\phantom{*}      \\ 
\hline

\end{tabular}
\end{sc}
\end{small}
\end{center}
\vskip -0.1in
\end{table*}

\section{Experiments} \label{sec:Experiments}

\subsection{Synthetic Data}

Our synthetic data experiments are based on those from \citet{Kuznetsov2020} in order to facilitate comparison with their method. Four distribution shift settings are considered: regularly occurring change points, gradual distribution drift, irregularly occurring change points and no distribution shift. Data are generated by the following equations for $t = 1, \ldots, 3000$ and $\epsilon_t \overset{\text{iid}}\sim N(0, 0.05^2)$:
\begin{itemize}
    \item[] \textbf{FixedRegime}: $Y_t = \theta_t Y_{t - 1} + \epsilon_t$, where  $\theta_t = -0.9$ if $t \in [1000, 2000]$ and 0.9 otherwise. \vspace{-0.2cm}
    \item[] \textbf{RandomWalk}: $Y_t = \theta_t Y_{t - 1} + \epsilon_t$, where $\theta_t = 1 - (t/1500)$. \vspace{-0.2cm}
    \item[] \textbf{RandomRegime}: $Y_t = \theta_{i(t)} Y_{t - 1} + \epsilon_t$, where $\theta_1=-0.5, \; \theta_2=0.9$ and $i_t$ is the stochastic process on $\{0, 1\}$ such that 
    \\$\mathbb{P}(i_{s + t} = i|i_{s + t -  1:s}=i, i_{s-1}\neq i) = (0.99998255)^t$. \vspace{-0.7cm}
    \item[] \textbf{Stat}: $Y_t = -0.5 Y_{t-1} + \epsilon_t$.
\end{itemize}
The task in each setting is to forecast the time series value $Y_t$ using a no-intercept linear model with a feature vector that consists of the last three values of the time series $X_t = \left(Y_{t-1}, Y_{t-2}, Y_{t-3}\right)$. The train-validation-test split for the synthetic data experiments is $D_{train} = \{(X_t, Y_t)\}_{t=1}^{2875}$, $D_{val} = \{(X_{t}, Y_{t})\}_{t=2876}^{2975}$ and $D_{test} = \{(X_t,Y_t)\}_{t=2976}^{3000}$. We use 192 Monte Carlo runs for each data setting.

\subsection{Real Data}

We consider a financial dataset consisting of 19 years of daily log-returns for 50 NYSE equities; see Appendix \ref{sec:Appendix - Data} for data processing details.

Our first real data experiment investigates distribution shift in the problem of risk factor modelling \cite{Fama1993}. Specifically, we linearly decompose each equity return $Y^{(i)}_t, \, i=1,\ldots,50$ in excess of the risk-free rate $\mathrm{RF}_t$ using a three factor model parameterised by $\theta_t^{(i)} \in \rn{4}$:
\begin{align*}
    Y^{(i)}_t - \mathrm{RF}_t = \theta_t^{(i)T}X_t
    \; \mathrm{\textit{where}} \; 
    X_t := \left(1, \mathrm{MR}_t, \mathrm{SB}_t, \mathrm{HL}_t\right)^T
\end{align*}
The three Fama-French factors correspond to market risk $(\mathrm{MR})$, the risk factor related to size $(\mathrm{SB})$ and the risk factor related to book-to-market equity  $(\mathrm{HL})$ \cite{Fama1993}. 

Our second real data experiment investigates distribution shift in the problem of forecasting the absolute values of financial returns \cite{Taylor1986}. For this experiment, we restrict our dataset to 15 Exchange Traded Fund (ETF) instruments tracking industry sectors, commodity and bond prices\footnote{The ETFs used are listed in the Appendix \ref{sec:Appendix - Data}.}. We forecast the next day absolute value of the return for each ETF, $Y^{(i)}_t, \, i=1,\ldots,15$, using the five previous absolute return values for that ETF with a linear model parameterised by $\theta_t^{(i)} \in \rn{5}$:
\begin{align*}
    | Y^{(i)}_t | = \theta_t^{(i)T}X_t   \; \mathrm{\textit{where}} \; X_t := (| Y^{(i)}_{t-1} |, \ldots, | Y^{(i)}_{t-5} |)^T. 
\end{align*}
For each real data time series, we use expanding time series cross validation to evaluate the methods. We start training for each time series using 6 years of data, the validation set is 150 days and we use the next 150 days as out-of-sample data. The train-validation-test split is moved forward by 150 days for the next model update.

\subsection{Methods} \label{sec:Experiments - Methods}
We consider two simple instances of forgetting mechanisms in our experiments
\begin{align} 
    \alpha(\tau; \eta) &= \exp(-\eta_1\tau) \label{eq:exp-decay}, \\
    \alpha(\tau; \eta) &= \exp(-\eta_1\tau - \eta_2 \tau^2 - \eta_3 \log(\tau + 1)). \label{eq:mixed-decay}
\end{align}
Forgetting mechanism (\ref{eq:exp-decay}) corresponds to exponential decay while (\ref{eq:mixed-decay}) corresponds to a mixture of various functional forms of decay. We fit each of these forgetting mechanisms using differentiable bi-level optimisation and call the resulting methods \texttt{GradExp} for mechanism (\ref{eq:exp-decay}) and \texttt{GradMixedDecay} for mechanism (\ref{eq:mixed-decay}). Our gradient optimisation routines are implemented using 5 restarts from random initialisation, with each run performing 50 epochs of stochastic gradient updates. As a comparison to gradient-based optimisation, we fit forgetting mechanism (\ref{eq:exp-decay}) using grid search over $\eta_1$: the corresponding method is known as \texttt{GridSearchExp}.  

\footnotetext[3]{Note that the independence assumption of the Wilcoxon test does not hold in the real data setting due to correlations between time series.}
We compare against five baselines: \texttt{Stationary}, \texttt{Window}, \texttt{StateSpace}, \texttt{ARIMA} and \texttt{DBF}. The \texttt{Stationary} method assumes no distribution shift and fits the underlying models using unweighted historical samples. The commonly used \texttt{Window} method applies uniform weights to a fixed length interval of the most recent samples. In the \texttt{StateSpace} model, the state is the vector of model parameters which is modelled using random-walk update state transition equations \cite{brockwell2009}. \texttt{ARIMA} is an autoregressive model (not applicable to the factor experiment) often used in econometrics \cite{brockwell2009}. \texttt{DBF} is an implementation of the two-step, discrepancy-based forecasting algorithm from \citet{Kuznetsov2020}. Details of model implementation and hyperparameters are reported in the Appendix \ref{sec:Appendix - Model Details}.

\subsection{Results}
We see from Table \ref{tab:results} that the \texttt{GradMixedDecay} performs best in 4 out of 6 datasets. It has near-best performance in the \texttt{RandomRegime} and \texttt{Stat} data settings. These results demonstrate the competitive predictive performance under distribution shift that can be achieved with the use of a forgetting mechanism which is trained using gradient-based bi-level optimisation. The outperformance of \texttt{GradMixedDecay} over 
\texttt{GradExp} illustrates the advantage of using more expressive differentiable forgetting mechanisms, which would be too computationally burdensome to fit using grid search. We see that \texttt{GradExp} has comparable performance to \texttt{GridSearchExp} on most datasets. This shows that gradient-based optimisation is capable of achieving comparable performance to grid search in the low-dimensional $\eta$ cases when grid search is feasible.

\section{Conclusion} \label{sec:Conclusion}

In summary, we propose learning forgetting mechanisms for time series prediction under distribution shift using gradient-based optimisation. Our approach is flexible, fast and can achieve competitive performance to baseline methods for time series prediction under distribution shift. In future work, we will investigate more complex forgetting mechanisms and sophisticated predictive models. Another interesting research extension would be to adapt our method to the online learning setting so that the predictive model can be updated in response to streaming data.
More broadly, our paper motivates further research between the topics of distribution shift, time series prediction and hyper-parameter learning. 

\section*{Acknowledgements}
The authors would like to thank Mihai Cucuringu, Gesine Reinert and Jake Topping for their feedback.
Both authors are supported by the EPSRC CDT in Modern Statistics and Statistical Machine Learning (EP/S023151/1) and the Alan Turing Institute (EP/N510129/1).

\bibliography{workshop_main}
\bibliographystyle{icml2022}

\newpage
\appendix
\onecolumn
\section{Data} \label{sec:Appendix - Data}
We consider the universe of NYSE equities spanning from 04-01-2000 to 31-12-2019 from Wharton's CRSP database \cite{WRDS}. The data consists of daily closing prices from which we compute daily log-returns. We also compute the average daily dollar volume that is traded for each equity. We subset to the 50 equities that have the largest average daily dollar volume. Any missing prices are forward-filled prior to the calculation of log-returns.

Factor returns were downloaded from Kenneth French's data repository 
\url{https://mba.tuck.dartmouth.edu/pages/faculty/ken.french/data_library.html}.

The ETF ticker names used for volatility experiments are $\{$`SPY', `IWM', `EEM', `TLT', `USO', `GLD', `XLF', `XLB', `XLK', `XLV', `XLI', `XLU', `XLY', `XLP', `XLE'$\}$.

\section{Model Details} \label{sec:Appendix - Model Details}
Each model is grid searched over its relevant hyper-parameters to maximise predictive performance on the validation set for each train-validation split of the data. In this Appendix, we provide details of the range of hyperparameter values searched over for each model.

Linear models in the \texttt{Stationary}, \texttt{Window}, \texttt{DBF}, \texttt{GridSearchExp}, \texttt{GradExp} and \texttt{GradMixedDecay} cases were fit using ridge regression with a squared error loss function, where the range of ridge penalty values searched over is $\lambda \in \{ 1 \times 10^{-i} : 3 \leq i \leq 6 \} \cup \{ 0 \}$. 
\begin{itemize}
    \item \textbf{\texttt{GradExp}, \texttt{GradMixedDecay}}:
    For optimising $\eta$ on the validation set we use standard mini-batch stochastic gradient descent with momentum for all experiments, keeping the learning rate and momentum constant at 0.1 and 0.9 respectively. We perform 5 restarts\footnote[4]{However, we have found that similar results can be obtained using a single run of gradient descent for most forgetting mechanisms and datasets.}
    of gradient descent from a random initialisation per run. We used a batch size of 32 examples, and train for 50 epochs per restart. We then select the value of $\eta$ with the lowest overall validation loss for each method.
    \item \textbf{\texttt{Window}}: 
    For a time series of length $T$, we grid search the length of the window over 25 values linearly spaced in $[5,T]$.
    \item \textbf{\texttt{GridSearchExp}}: 
    We grid search $\eta_1$ in Equation \eqref{eq:exp-decay} over 25 values. In order to ensure that the grid of values considered is of plausible a priori range, the $\eta_1$ grid is computed based on the window length grid which was searched over in the \texttt{Window} method. In particular, for each window length value considered, we select a value of $\eta_1$ that results in the $\texttt{GridSearchExp}$ forgetting mechanism assigning almost all of its weight to the same span length.
    \item \textbf{\texttt{ARIMA}}: 
    The hyperparameters of \texttt{ARIMA} are the autoregressive lag $p$, the differencing order $d$ and the number of moving average components $q$. Following \citet{Kuznetsov2020}, we grid search over $p, d, q \in \{ 0, 1, 2 \}^3$. The model is fit using maximum likelihood under a Gaussian emission model \cite{brockwell2009}
    \item \textbf{\texttt{StateSpace}}:
    The state-space model's observation equation is a linear model with Gaussian noise where the states correspond to the predictive model's regression coefficients for the given problem. In the state equation, the states are updated with a Gaussian increment. The model is fit using maximum likelihood.
    \item \textbf{\texttt{DBF}}: The \texttt{DBF} algorithm is an iterative two stage procedure that iteratively fits the predictive model and the weights in the ERM problem (see \citet{Kuznetsov2020}, Section 7, Equation 14). Their procedure is run for 5 iterations. The lookback parameter, $s$, corresponding to the window in which the \textit{discrepancy} is thought to be bounded \cite{Kuznetsov2020} is set to the value used in the original paper $s=20$. The \texttt{DBF} algorithm has two penalty parameters $\lambda_1$ and $\lambda_2$, which correspond to regularisation on the weighting scheme and the model parameters. We grid search both of these parameters over the same grid as the ridge regression penalty. 
\end{itemize}

\section{Bi-Level Optimisation for Deep Learning} \label{sec: Appendix - Bi-Level Optimisation for Deep Learning}
In ongoing work not presented in this paper, we found that further considerations are required when the lower level problem has multiple local minima (as is the case when using a neural network as the predictive model).
We have empirically found that flatness-aware learning \cite{SWA} of the lower level problem is important for the stability and generalisation performance of the bi-level optimisation solution. We conjecture that this is because a sharp minima leads to a large Hessian condition number and therefore unstable inversion in Equation \eqref{eq:theta_alpha_deriv_alpha}. 

Secondly, we have observed that identity approximation to the Hessian in \cref{eq:theta_alpha_deriv_alpha} yields competitive results while reducing the computational cost of running bi-level optimisation using a highly-parameterised predictive model such as a neural network. A number of other Hessian approximation schemes have been proposed in the gradient hyper-parameter optimisation literature, see \citet{duv_hyp} for a comparison of these approaches in the context of hyper-parameter optimisation. 


\end{document}